\documentclass[runningheads]{llncs}
\usepackage[numbers,square]{natbib}
\usepackage{orcidlink}
\usepackage{graphicx}
\usepackage{bold-extra}
\usepackage{xcolor}
\definecolor{codegreen}{rgb}{0,0.6,0}
\definecolor{codegray}{rgb}{0.5,0.5,0.5}
\definecolor{codepurple}{rgb}{0.58,0,0.82}
\definecolor{backcolour}{rgb}{0.98,0.98,0.96}
\usepackage{listings}
\lstdefinestyle{mystyle}{
    backgroundcolor=\color{backcolour},   
    commentstyle=\color{codegreen},
    keywordstyle=\color{magenta},
    numberstyle=\tiny\color{codegray},
    stringstyle=\color{codepurple},
    basicstyle=\ttfamily\scriptsize,
    breakatwhitespace=false,         
    breaklines=true,                 
    captionpos=b,                    
    keepspaces=true,                 
    numbers=none,                    
    numbersep=5pt,                  
    showspaces=false,                
    showstringspaces=false,
    showtabs=false,                  
    tabsize=2
}
\usepackage{array}

\newcommand{\tb}[1]{#1}

\begin{document}

\title{
A Melting Pot of Evolution and Learning\thanks{This research was partially supported by the following grants:
Israeli Innovation Authority through the Trust.AI consortium;
Israeli Science Foundation grant no. 2714/19;
Israeli Smart Transportation Research Center (ISTRC);
Israeli Council for Higher Education (CHE) via the Data Science Research Center, Ben-Gurion University of the Negev, Israel.
}\\{\scriptsize (To Appear in \textit{Proceedings of Genetic Programming Theory \& Practice XX, 2023)}}}

\titlerunning{A Melting Pot of Evolution and Learning}

\author{
Moshe Sipper\,\orcidlink{0000-0003-1811-472X}\inst{1} \and
Achiya Elyasaf\,\orcidlink{0000-0002-4009-5353}\inst{2} \and
Tomer Halperin\inst{1} \and
Zvika Haramaty\,\orcidlink{0000-0002-6225-188X}\inst{1} \and
Raz Lapid\,\orcidlink{0000-0002-4818-9338}\inst{1,3} \and
Eyal Segal\inst{1} \and
Itai Tzruia\inst{1} \and
Snir Vitrack Tamam\inst{1}}

\authorrunning{M. Sipper et al.}

\institute{Department of Computer Science, Ben-Gurion University of the Negev, Beer-Sheva, 8410501, Israel
\and
Department of Software and Information Systems Engineering, Ben-Gurion University of the Negev, Beer-Sheva, 8410501, Israel
\and
DeepKeep, Tel-Aviv, Israel \\
\email{sipper@bgu.ac.il}\\
\url{http://www.moshesipper.com/}}

\maketitle

\begin{abstract}
We survey eight recent works by our group, involving the successful blending of evolutionary algorithms with machine learning and deep learning:
\begin{enumerate}
\item Binary and Multinomial Classification through Evolutionary Symbolic Regression,
\item Classy Ensemble: A Novel Ensemble Algorithm for Classification,
\item EC-KitY: Evolutionary Computation Tool Kit in Python,
\item Evolution of Activation Functions for Deep Learning-Based Image Classification,
\item Adaptive Combination of a Genetic Algorithm and Novelty Search for Deep Neuroevolution,
\item An Evolutionary, Gradient-Free, Query-Efficient, Black-Box Algorithm for Generating Adversarial Instances in Deep Networks,
\item Foiling Explanations in Deep Neural Networks,
\item Patch of Invisibility: Naturalistic Black-Box Adversarial Attacks on Object Detectors.
\end{enumerate}

\end{abstract}

\section{Introduction}
\label{sec:intro}
In Evolutionary Computation (EC)---or Evolutionary Algorithms (EAs)---core concepts from evolutionary biology—inheritance, random variation, and selection—are harnessed in algorithms that are applied to complex computational problems. As discussed by \citet{Sipper2017ec}, EAs present several important benefits over popular machine learning (ML) methods, including: less reliance on the existence of a known or discoverable gradient within the search space; ability to handle design problems, where the objective is to design new entities from scratch; fewer required a priori assumptions about the problem at hand; seamless integration of human expert knowledge; ability to solve problems where human expertise is very limited; support of interpretable solution representations; support of multiple objectives. 

Importantly, these strengths often dovetail with weak points of ML algorithms, which has resulted in an increasing number of works that fruitfully combine the fields of EC with ML or deep learning (DL). Herein, we will survey eight recent works by our group, which are at the intersection of EC, ML, and DL:
\begin{itemize}
    \item Machine Learning (Section~\ref{sec:ml})
        \begin{enumerate}
            \item Binary and Multinomial Classification through Evolutionary Symbolic Regression \cite{Sipper2022esr} (Section~\ref{sec:multinomial})
            \item Classy Ensemble: A Novel Ensemble Algorithm for Classification \cite{sipper2022classy} (Section~\ref{sec:classy})
            \item EC-KitY: Evolutionary Computation Tool Kit in Python \cite{eckity2023} (Section~\ref{sec:eckity})
        \end{enumerate}
    \item Deep Learning (Section~\ref{sec:dl})
        \begin{enumerate}
            \setcounter{enumi}{3}
            \item Evolution of Activation Functions for Deep Learning-Based Image Classification \cite{Lapid2022} (Section~\ref{sec:af})
            \item Adaptive Combination of a Genetic Algorithm and Novelty Search for Deep Neuroevolution \cite{SegalS22} (Section~\ref{sec:novelty})
        \end{enumerate}
    \item Adversarial Deep Learning (Section~\ref{sec:adv})
        \begin{enumerate}
            \item An Evolutionary, Gradient-Free, Query-Efficient, Black-Box Algorithm for Generating Adversarial Instances in Deep Networks \cite{Lapid2022Query} (Section~\ref{sec:query})
            \item Foiling Explanations in Deep Neural Networks \cite{Vitrack2023} (Section~\ref{sec:foil})
            \item Patch of Invisibility: Naturalistic Black-Box Adversarial Attacks on Object Detectors \cite{Lapid2023} (Section~\ref{sec:patch})
        \end{enumerate}
\end{itemize}

If one's interest is piqued by a particular project we invite them to peruse the respective, cited, full paper (which are all freely available online).

\section{Machine Learning}
\label{sec:ml}

\subsection{Binary and Multinomial Classification through Evolutionary Symbolic Regression \cite{Sipper2022esr}}
\label{sec:multinomial}
Classification is an important subfield of supervised learning. As such, many powerful algorithms have been designed over the years to tackle both binary datasets as well as multinomial, or multiclass ones. Symbolic regression (SR) is a family of algorithms that aims to find regressors of arbitrary complexity. \citet{Sipper2022esr} showed that evolutionary SR-based \textit{regressors} can be successfully converted into performant \textit{classifiers}.

We devised and tested three evolutionary SR-based classifiers: GPLearnClf, CartesianClf, and ClaSyCo. The first two are based on the one-vs-rest approach, while the last one is inherently multinomial. 

\textit{GPLearnClf} is based on the GPLearn package \cite{stephens2019gplearn}, which implements tree-based Genetic Programming (GP) symbolic regression, is relatively fast, and---importantly---interfaces seamlessly with Scikit-learn \cite{scikit-learn}.
GPLearnClf evolves $C$ separate populations independently, each fitted to a specific class by considering as target values the respective column vector (of $C$ column vectors) of the one-hot-encoded target vector $y$. The fitness function is based on log loss (aka binary cross-entropy). Prediction is carried out by outputting the argmax of the set of best evolved individuals (one from each population). The hyperparameters to tune were population size and generation count. 

\textit{CartesianClf} is based on Cartesian GP (CGP), which grew from a method of evolving digital circuits \cite{miller2011cartesian}. It is called `Cartesian' because it represents a program using a two-dimensional grid of nodes. 
The CGP package we used \cite{Ohjeah} evolves the population in a $(1 + \lambda)$-manner, i.e., in each generation it creates $\lambda$ offspring (we used the default $\lambda =4$) and compares their fitness to the parent individual. The fittest individual carries over to the next generation; in case of a draw, the offspring is preferred over the parent. Tournament selection is used (tournament size $=|population|$), single-point mutation, and no crossover. 

We implemented CartesianClf similarly to GPLearnClf in a one-vs-rest manner, with $C$ separate populations evolving independently, using binary cross-entropy as fitness.
The hyperparameters to tune were number of rows, number of columns, and maximum number of generations.

\textit{ClaSyCo} (\textbf{Cla}ssification through \textbf{Sy}mbolic Regression and\linebreak \textbf{Co}evolution) also employs $C$ populations of trees;
however, these are not evolved independently as with the one-vs-rest method (as done with GPLearnClf and CartesianClf)---but in tandem through \textit{cooperative coevolution}.

A cooperative coevolutionary algorithm involves a number of evolving populations, which come together to obtain problem solutions. The fitness of an individual in a particular population depends on its ability to collaborate with individuals from the other populations \cite{Pena2001,sipper2019omnirep}.

Specifically, in our case,
an individual SR tree $i$ in population $c$, $\mathit{gp}^c_i$, $i \in \mathit{\{1,\ldots,n\_pop\}}$, $c \in \{1,\ldots,C\}$, is assigned fitness through the following steps (we describe this per single dataset sample, although in practice fitness computation is vectorized by Python):
\begin{enumerate}
    \item Individual $\mathit{gp}^c_i$ computes an output $\hat{y}^c_i$ for the sample under consideration.
    
    \item Obtain the best-fitness classifier of the previous generation, $\mathit{gp}^{c'}_{\mathit{best}}$, for each population $c'$, $c' \in \{1,\ldots,C\}$, $c' \neq c$
    (these are called ``representatives'' or ``cooperators'' \cite{Pena2001}).
    
    \item Each $\mathit{gp}^{c'}_{\mathit{best}}$ computes an output $\hat{y}^{c'}_{\mathit{best}}$ for the sample under consideration.
    
    \item We now have $C$ output values, $\hat{y}^{1}_{\mathit{best}}, ... , \hat{y}^c_i , ... , \hat{y}^{C}_{\mathit{best}}$. 
    
    \item Compute $\sigma(\hat{y}^{1}_{\mathit{best}}, ... , \hat{y}^c_i , ... , \hat{y}^{C}_{\mathit{best}})$, where $\sigma$ is the softmax function.
    
    \item Assign a fitness score to $\mathit{gp}^c_i$ using the cross-entropy loss function. (NB: only individual $\mathit{gp}^c_i$ is assigned fitness---all other $C-1$ individuals are representatives.)
\end{enumerate}

Tested over 162 datasets and compared to three state-of-the-art machine learning algorithms---XGBoost, LightGBM, and a deep neural network---we found our algorithms to be competitive.
Further, we demonstrated how to find the best method for one's dataset automatically, through the use of Optuna, a state-of-the-art hyperparameter optimizer \cite{akiba2019optuna}.

\subsection{Classy Ensemble: A Novel Ensemble Algorithm for Classification \cite{sipper2022classy}}
\label{sec:classy}
\citet{sipper2022classy} presented \textit{Classy Ensemble}, a novel ensemble-generation algorithm for classification tasks, which aggregates models through a weighted combination of per-class accuracy. 

The field of ensemble learning has an illustrious history spanning several decades. Indeed, we ourselves employed this paradigm successfully in several recent works:
\begin{itemize}
    \item \citet{Sipper2020,Sipper2021cml} presented \textit{conservation machine learning}, which conserves models across runs, users, and experiments. As part of this work we compared multiple ensemble-generation methods, also introducing \textit{lexigarden}---which is based on lexicase selection, a performant selection technique for evolutionary algorithms \cite{metevier2019lexicase,spector2012assessment}.

    \item \citet{sipper2021symbolic} presented SyRBo---Symbolic-Regression Boosting---an ensemble method based on strong learners that are combined via boosting, used to solve regression tasks.

    \item \citet{Sipper2021addgBoost} introduced AddGBoost, a gradient boosting-style algorithm, wherein the decision tree is replaced by a succession of stronger learners, which are optimized via a state-of-the-art hyperparameter optimizer.

    \item \citet{Sipper2023Combining} presented a comprehensive, stacking-based framework for combining deep learning with good old-fashioned machine learning, called Deep GOld. The framework involves ensemble selection from 51 retrained pretrained deep networks as first-level models, and 10 ML algorithms as second-level models.
\end{itemize}

Classy Ensemble receives as input a collection of fitted models, each one's overall accuracy score, and per-class accuracy, i.e., each model's accuracy values as computed separately for every class (note: we used \texttt{scikit-learn}'s \cite{scikit-learn} \texttt{balanced\_accuracy\_score}, which avoids inflated performance estimates on imbalanced datasets).

Classy Ensemble adds to the ensemble the \textit{topk} best-performing (over validation set) models, \textit{for each class}. A model may be in the \textit{topk} set for more than one class. Thus, for each model in the ensemble, we also maintain a list of classes for which it is a \textit{voter}, i.e., its output for each voter class is taken into account in the final aggregation. 
The binary voter vector of size \textit{n\_classes} is set to 1 for classes the model is permitted to vote for, 0 otherwise.

Thus, a model not in the ensemble is obviously not part of the final aggregrated prediction; further, a model \textit{in} the ensemble is only ``allowed'' to vote for those classes for which it is a voter---i.e., for classes it was amongst the \textit{topk}.

Classy Ensemble provides a prediction by aggregating its members' predicted-class probabilities, weighted by the overall validation score, and taking into account voter permissions.

Tested over 153 machine learning datasets we demonstrated that Classy Ensemble outperforms two other well-known aggregation algorithms---order-based pruning and clustering-based pruning---as well as our aforementioned lexigarden ensemble generator. 

We then enhanced Classy Ensemble with a genetic algorithm, creating \textit{Classy Evolutionary Ensemble}, wherein an evolutionary algorithm is used to select the set of models which Classy Ensemble picks from. This latter algorithm was able to improve state-of-the-art deep learning models over the well-known, difficult ImageNet dataset.

\subsection{EC-KitY: Evolutionary Computation Tool Kit in Python \cite{eckity2023}}
\label{sec:eckity}
There is a growing community of researchers and practitioners who combine evolution and learning. Having used several EC open-source software packages over the years we identified a large ``hole'' in the software landscape---there was a lacuna in the form of an EC package that is:

\begin{enumerate}
\item A comprehensive toolkit for running evolutionary algorithms.
\item Written in Python.
\item Can work with or without \tb{scikit-learn} (aka \tb{sklearn}), the most popular ML library for Python. To wit, the package should support both sklearn and standalone (non-sklearn) modes.
\item Designed with modern software engineering in mind.
\item Designed to support all popular EC paradigms: genetic algorithms (GAs), genetic programming (GP), evolution strategies (ES), coevolution, multi-objective, etc'.
\end{enumerate}

While there are several EC Python packages, none fulfill \textit{all} five requirements. Some are not written in Python, some are badly documented, some do not support multiple EC paradigms, and so forth. 
Importantly for the ML community, most tools do not intermesh with extant ML tools. Indeed, we have personally had experience with the hardships of combining EC tools with scikit-learn when doing evolutionary machine learning. 

Thus was born \tb{EC-KitY}: a comprehensive Python library for doing EC, licensed under the BSD 3-Clause License, and compatible with \tb{scikit-learn}. Designed with modern software engineering and machine learning integration in mind, \tb{EC-KitY} can support all popular EC paradigms, including genetic algorithms, genetic programming, coevolution, evolutionary multi-objective optimization, and more. 

\tb{EC-KitY} can work both in standalone, non-sklearn mode, and in sklearn mode.
Below we show two code examples that solve a symbolic regression problem.

In standalone mode the user can run an EA with a mere three lines of code:
\begin{lstlisting}[language=Python,upquote=true]
from eckity.algorithms.simple_evolution import SimpleEvolution
from eckity.subpopulation import Subpopulation
from examples.treegp.non_sklearn_mode.symbolic_regression.sym_reg_evaluator import SymbolicRegressionEvaluator

algo = SimpleEvolution(Subpopulation(SymbolicRegressionEvaluator()))
algo.evolve()
print('algo.execute(x=2, y=3, z=4):', algo.execute(x=2, y=3, z=4))
\end{lstlisting}

Running an EA in sklearn mode is just as simple:
\begin{lstlisting}[language=Python,upquote=true]
from sklearn.datasets import make_regression
from sklearn.metrics import mean_absolute_error
from sklearn.model_selection import train_test_split

from eckity.algorithms.simple_evolution import SimpleEvolution
from eckity.creators.gp_creators.full import FullCreator
from eckity.genetic_encodings.gp.tree.utils import create_terminal_set
from eckity.sklearn_compatible.regression_evaluator import RegressionEvaluator
from eckity.sklearn_compatible.sk_regressor import SKRegressor
from eckity.subpopulation import Subpopulation

X, y = make_regression(n_samples=100, n_features=3)
terminal_set = create_terminal_set(X)
algo = SimpleEvolution(
         Subpopulation(creators=FullCreator(terminal_set=terminal_set),
                       evaluator=RegressionEvaluator()))
regressor = SKRegressor(algo)
X_train, X_test, y_train, y_test = train_test_split(X, y, test_size=0.2)
regressor.fit(X_train, y_train)
print('MAE on test set:', 
      mean_absolute_error(y_test, regressor.predict(X_test)))
\end{lstlisting}

We recently taught a course in which 48 students worked in groups of two or three, submitting a total of 22 projects that used \tb{EC-KitY} to solve a diverse array of complex problems, including evolving Flappy Bird agents, evolving blackjack strategies, evolving Super Mario agents, evolving chess players, and solving problems such as maximum clique and vehicle routing. \tb{EC-KitY} proved quite up to the tasks.

\section{Deep Learning}
\label{sec:dl}

\subsection{Evolution of Activation Functions for Deep Learning-Based Image Classification \cite{Lapid2022}}
\label{sec:af}
Artifical Neural Networks (ANNs), and, specifically, Deep Neural Networks\linebreak (DNNs), have gained much traction in recent years and are now being effectively put to use in a variety of applications. Considerable work has been done to improve training and testing performance, including various initialization techniques, weight-tuning algorithms, different architectures, and more. However, one hyperparameter is usually left untouched: the activation function (AF). While recent work has seen the design of novel AFs \cite{agostinelli2014learning,saha2019evolution,sharma2017activation,sipper2021neural}, the Rectified Linear Unit (ReLU) remains by far the most commonly used one, mainly due to its overcoming the vanishing-gradient problem, thus affording faster learning and better performance.

\citet{Lapid2022} introduced a novel coevolutionary algorithm to evolve AFs for image-classification tasks. Our method is able to handle the simultaneous coevolution of three types of AFs: input-layer AFs, hidden-layer AFs, and output-layer AFs. We surmised that combining different AFs throughout the architecture may improve the network's performance. 

We devised a number of evolutionary algorithms, including a coevolutionary one,
comprising three separate populations: 1) input-layer AFs, 2) hidden-layer AFs, 3) output-layer AFs. Combining three individuals---one from each population---results in an AF architecture that can be evaluated.

We compared our novel algorithm to four different methods: standard ReLU- or LeakyReLU-based networks, networks whose AFs are produced randomly, and two forms of single-population evolution, differing in whether an individual represents a single AF or three AFs. We chose ReLU and LeakyReLU as baseline AFs since we noticed that they are the most-used functions in the deep-learning domain.

\begin{figure}
    \centering
    \includegraphics[width=0.95\textwidth]{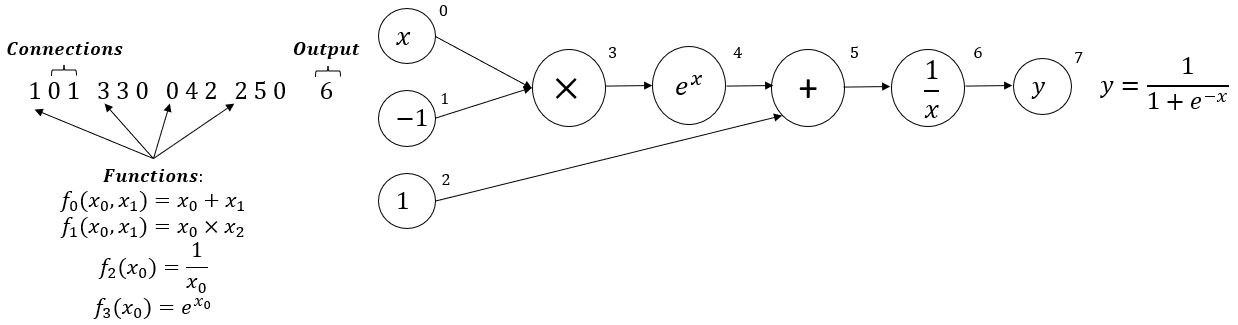}
    \caption{A sample CGP individual, with 3 inputs---$x, -1, 1$---and 1 output---$y$. 
    The genome consists of 5 3-valued genes, per 4 functional units, plus the output specification (no genes for the inputs). 
    The first value of each 3-valued gene is the function's index in the lookup table of functions (bottom-left), and the remaining two values are parameter nodes. The last gene determines the outputs to return. 
    In the example above, with $n_i$ representing node $i$: 
    node 3, gene 101, $f_1(n_0, n_1)=n_0 \times n_1$;
    node 4, gene 330, $f_3(n_3)=e^{n_3}$ (unary function, 3rd gene value ignored);
    node 5, gene 042, $f_0(n_4,n_2)=n_4 + n_2$; 
    node 6, gene 250, $f_2(n_5)=\frac{1}{n_5}$;
    node 7, output node, $n_6$ is the designated output value.
    The topology is fixed throughout evolution, while the genome evolves.}
    \label{fig:ind_example}
\end{figure}

We used Cartesian genetic programming (CGP), wherein an evolving individual is represented as a two-dimensional grid of computational nodes---often an a-cyclic graph---which together express a program \cite{10.1145/1388969.1389075}. An individual is represented by a linear genome, composed of integer genes, each encoding a single node in the graph, which represents a specific function. A node consists of a function, from a given table of functions, and connections, specifying  where the data for the node comes from.
A sample individual in the evolving CGP population, representing the well-known sigmoid AF, is shown in Figure~\ref{fig:ind_example}.

Tested on four datasets---MNIST, FashionMNIST, KMNIST, and USPS---coevolution proved to be a performant algorithm for finding good AFs and AF architectures.

\subsection{Adaptive Combination of a Genetic Algorithm and Novelty Search for Deep Neuroevolution \cite{SegalS22}}
\label{sec:novelty}
As the field of Reinforcement Learning (RL) \cite{sutton2018reinforcement} is being applied to harder tasks, two unfortunate trends emerge: larger policies that require more computing time to train, and ``deceptive'' optima. While gradient-based methods do not scale well to large clusters, evolutionary computation (EC) techniques have been shown to greatly reduce training time by using modern distributed infrastructure \cite{arxiv.1703.03864,arxiv.1712.06567}.

The problem of deceptive optima has long since been known in the EC community: Exploiting the objective function too early might lead to a sub-optimal solution, and attempting to escape it incurs an initial loss in the objective function. Novelty Search (NS) mitigates this issue by ignoring the objective function while searching for new behaviors \cite{Lehman2008}. This method had been shown to work for RL \cite{arxiv.1712.06567}. 

While both genetic algorithms (GAs) and NS have been shown to work in different environments \cite{arxiv.1712.06567}, we attempted in \cite{SegalS22} to combine the two to produce a new algorithm that does not fall behind either, and in some scenarios surpasses both. 

\citet{SegalS22} proposed a new algorithm: Explore-Exploit $\gamma$-Adaptive Learner ($E^2\gamma AL$, or EyAL). By preserving a dynamically-sized niche of novelty-seeking agents, the algorithm manages to maintain population diversity, exploiting the reward signal when possible and exploring otherwise. The algorithm combines both the exploitative power of a GA and the explorative power of NS, while maintaining their simplicity and elegance. 

Our experiments showed that EyAL outperforms NS in most scenarios, while being on par with a GA---and in some scenarios it can outperform both. EyAL also allows the substitution of the exploiting component (GA) and the exploring component (NS) with other algorithms, e.g., Evolution Strategy and Surprise Search, thus opening the door for future research.

\section{Adversarial Deep Learning}
\label{sec:adv}

\subsection{An Evolutionary, Gradient-Free, Query-Efficient, Black-Box Algorithm for Generating Adversarial Instances in Deep Networks \cite{Lapid2022Query}}
\label{sec:query}
Despite their success, recent studies have shown that DNNs are vulnerable to adversarial attacks. A barely detectable change in an image can cause a misclassification in a well-trained DNN. Targeted adversarial examples can even evoke a misclassification of a specific class (e.g., misclassify a car as a cat). Researchers have demonstrated that adversarial attacks are successful in the real world and may be produced for data modalities beyond imaging, e.g., natural language and voice recognition \cite{wang2019natural,morris2020textattack,carlini2018audio,schonherr2018adversarial}. DNNs' vulnerability to adversarial attacks has raised concerns about applying these techniques to safety-critical applications.

To discover effective adversarial instances, most past work on adversarial attacks has employed gradient-based optimization \cite{goodfellow2014explaining,papernot2016limitations,carlini2017towards,gu2014towards,moosavi2016deepfool}. Gradient computation can only be executed if the attacker is fully aware of the model architecture and weights. Thus, these approaches are only useful in a white-box scenario, where an attacker has complete access and control over a targeted DNN. Attacking real-world AI systems, however, might be far more arduous. The attacker must consider the difficulty of implementing adversarial instances in a black-box setting, in which no information about the network design, parameters, or training data is provided---the attacker is exposed only to the classifier's input-output pairs. In this context, a typical strategy has been to attack trained replacement networks and hope that the generated examples transfer to the target model \cite{papernot2017practical}. The substantial mismatch between the alternative model and the target model, as well as the significant computational cost of alternative network training, often renders this technique ineffective.

\citet{Lapid2022Query} assumed a real-world, black-box attack scenario, wherein a DNN's input and output may be accessed but not its internal configuration. We focused on a scenario in which a specific DNN is an image classifier, specifically, a convolutional neural network (CNN), which accepts an image as input and outputs a probability score for each class.

We presented \textit{QuEry Attack} (for \textbf{Qu}ery-Efficient \textbf{E}volutiona\textbf{ry} \textbf{Attack}):
an evolutionary, gradient-free optimization approach for generating adversarial instances, more suitable for real-life scenarios, because usually there is no access to a model's internals, including the gradients; thus, it is important to craft attacks that do not use gradients. Our proposed attack can deal with either constrained ($\epsilon$ value that constrains the norm of the allowed perturbation) or unconstrained (no constraint on the norm of the perturbation) problems, and focuses on constrained, untargeted attacks. We believe that our framework can be easily adapted to the targeted setting.

\begin{figure}
\centering
\begin{tabular}{ccc}
Original image & Successful attack & Failed attack \\[5pt]
\includegraphics[width=0.13\textwidth]{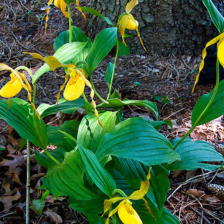} & 
\includegraphics[width=0.13\textwidth]{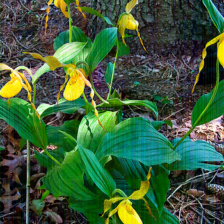} &
\includegraphics[width=0.13\textwidth]{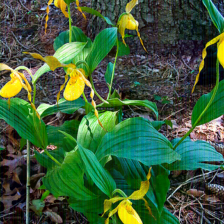} \\[5pt]

\includegraphics[width=0.13\textwidth]{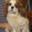} &
\includegraphics[width=0.13\textwidth]{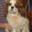} &
\includegraphics[width=0.13\textwidth]{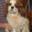} \\[5pt]

\includegraphics[width=0.13\textwidth]{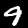} &
\includegraphics[width=0.13\textwidth]{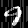} &
\includegraphics[width=0.13\textwidth]{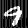} \\
\end{tabular}
\caption{Examples of adversarial attacks generated by QuEry Attack. With higher resolution the attack becomes less visible to the naked eye. The differences between images that successfully attack the model and those that do not are subtle.
         Top row: Imagenet ($l_\infty = 6/255$). 
         Middle row: CIFAR10 ($l_\infty = 6/255$).
         Bottom row: MNIST ($l_\infty = 60/255$).
         Left: the original image.
         Middle: a successful attack.
         Right: A failed attack.}
\label{fig:attacks}
\end{figure}

Figure~\ref{fig:attacks} shows examples of successful and unsuccessful instances of images generated by QuEry Attack, evaluated against ImageNet, CIFAR10, and MNIST.

QuEry Attack is a strong and fast attack that employs a gradient-free optimization strategy. We tested QuEry Attack against MNIST, CIFAR10, and ImageNet models, comparing it to other commonly used algorithms. We evaluated QuEry Attack's performance against non-differential transformations and robust models, and it proved to succeed in both scenarios.

\subsection{Foiling Explanations in Deep Neural Networks \cite{Vitrack2023}}
\label{sec:foil}
In order to render a DL model more interpretable, various explainable algorithms have been conceived. \citet{van2004explainable} coined the term \textit{Explainable Artificial Intelligence} (XAI), which refers to AI systems that ``can explain their behavior either during execution or after the fact''. In-depth research into XAI methods has been sparked by the success of Machine Learning systems, particularly Deep Learning, in a variety of domains, and the difficulty in intuitively understanding the outputs of complex models, namely, how did a DL model arrive at a specific decision for a given input.

Explanation techniques have drawn increased interest in recent years due to their potential to reveal hidden properties of DNNs \cite{dovsilovic2018explainable}. For safety-critical applications, interpretability is essential, and sometimes even legally required.

The importance assigned to each input feature for the overall classification result may be observed through explanation maps, which can be used to offer explanations. Such maps can be used to create defenses and detectors for adversarial attacks \cite{walia2022using,fidel2020explainability,kao2022rectifying}.

\citet{Vitrack2023} showed that these explanation maps can be transformed into any target map, using only the maps and the network's output probability vector. This was accomplished by adding a perturbation to the input that is scarcely (if at all) noticeable to the human eye. This perturbation has minimal effect on the neural network's output, therefore, in addition to the classification outcome, the probability vector of all classes remains virtually identical.

\begin{figure}
    \centering
    \includegraphics[width=0.95\textwidth]{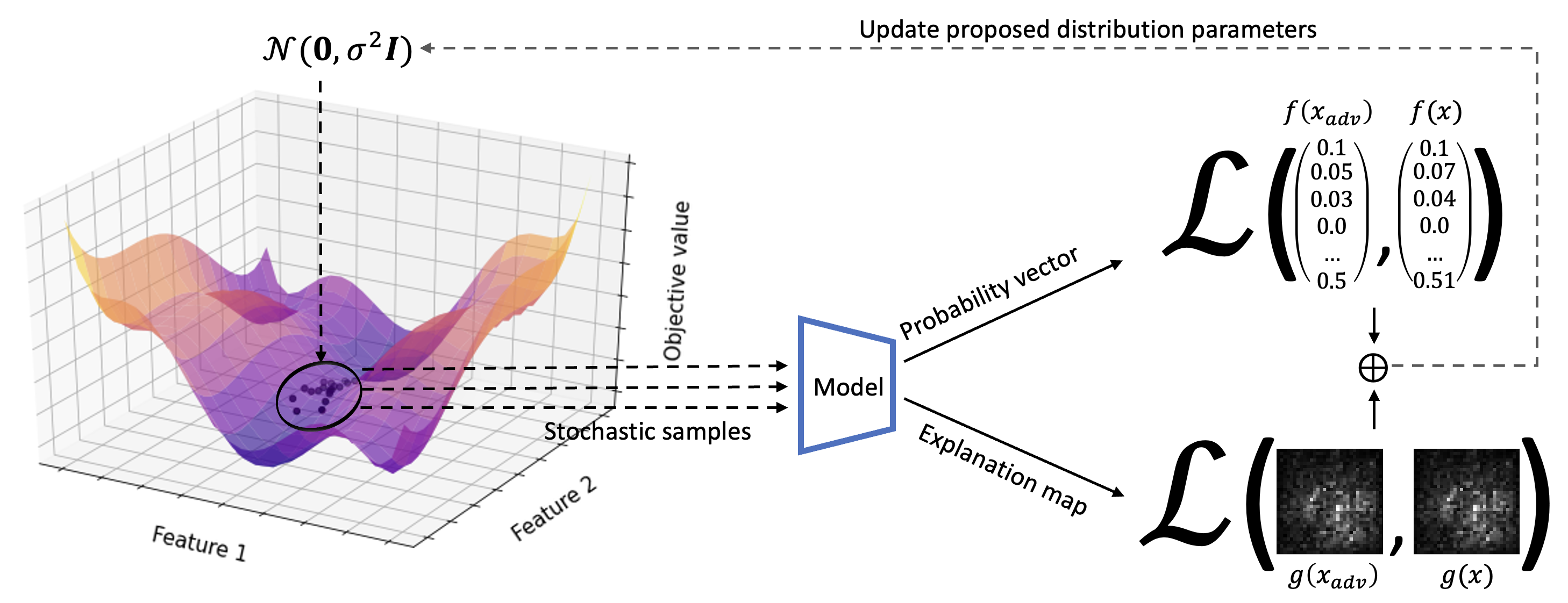}
    \caption{Schematic of AttaXAI. Individual images are sampled from the population's distribution $\mathcal{N}(\mu, \sigma)$, and fed into the model (Feature 1 and Feature 2 are image features, e.g., two pixel values; in reality the dimensionality is much higher). Then, the fitness function, i.e. the loss, is calculated using the output probability vectors and the explanation maps to approximate the gradient and update the distribution parameters, $\mu$ and $\sigma$.}
    \label{fig:AttaXAI}
\end{figure}

Our black-box algorithm, AttaXAI, enables manipulation of an image through a barely noticeable perturbation, without the use of any model internals, such that the explanation fits any given target explanation. AttaXAI explores the space of images through evolution, ultimately producing an adversarial image; it does so by continually updating a Gaussian probability distribution, used to sample the space of perturbations. By continually improving this distribution the search improves (Figure~\ref{fig:AttaXAI}).

\begin{figure}
\centering
\small
\setlength\tabcolsep{0pt}
\begin{tabular}{>{\centering}m{0.14\textwidth} >{\centering}m{0.14\textwidth} >{\centering}m{0.14\textwidth} >{\centering}m{0.14\textwidth} >{\centering}m{0.14\textwidth} >{\centering\arraybackslash}m{0.14\textwidth}}
 
$x$ & $x_{target}$ & $x_{adv}$ & $g(x)$ & $g(x_{target})$ & $g(x_{adv})$ \\

\includegraphics[width = 0.14\textwidth, trim={3cm 2cm 3cm 1cm}, keepaspectratio]{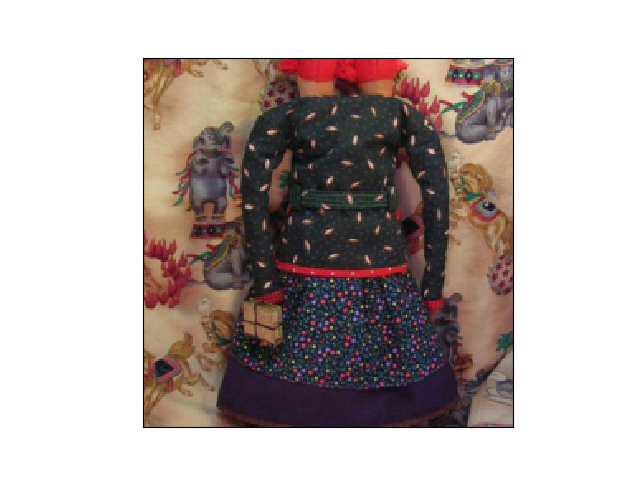} &
\includegraphics[width = 0.14\textwidth, trim={3cm 2cm 3cm 1cm}, keepaspectratio]{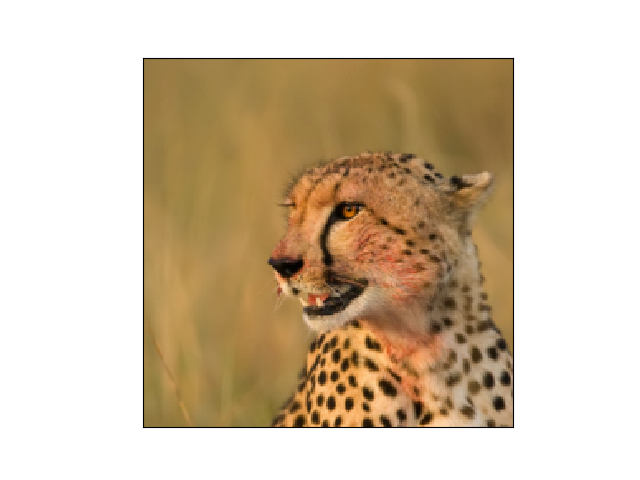} &
\includegraphics[width = 0.14\textwidth, trim={3cm 2cm 3cm 1cm}, keepaspectratio]{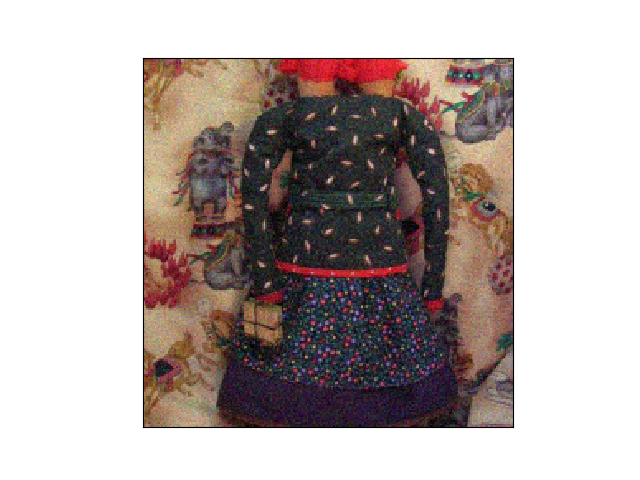} &
\includegraphics[width = 0.14\textwidth, trim={3cm 2cm 3cm 1cm}, keepaspectratio]{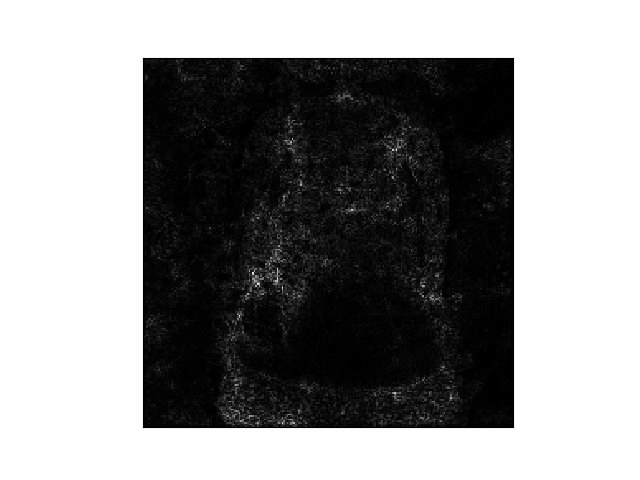} &
\includegraphics[width = 0.14\textwidth, trim={3cm 2cm 3cm 1cm}, keepaspectratio]{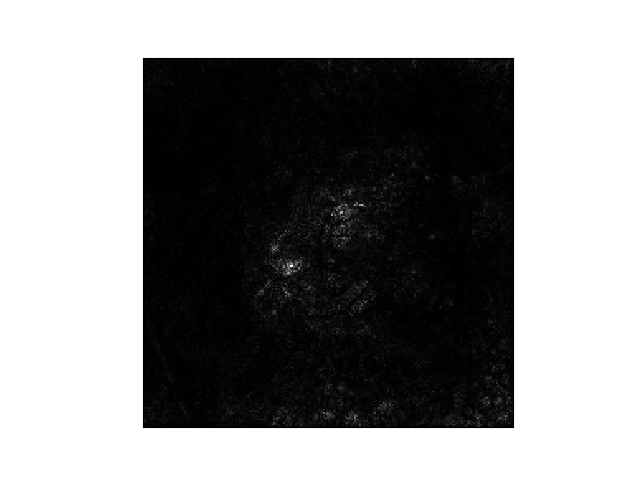} &
\includegraphics[width = 0.14\textwidth, trim={3cm 2cm 3cm 1cm}, keepaspectratio]{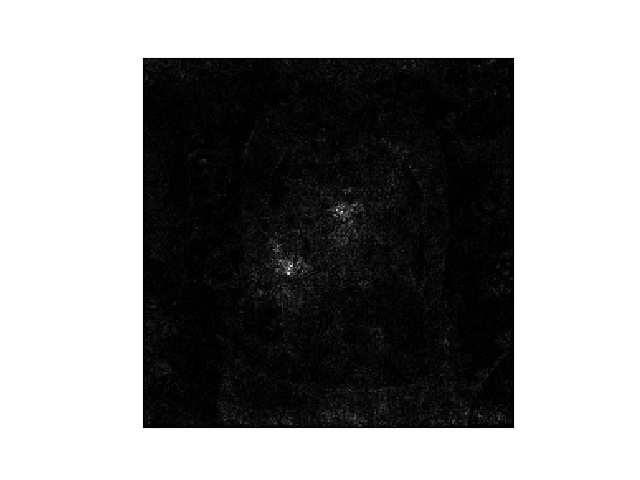} 
\\
\end{tabular}
\caption{An attack generated by AttaXAI.
Dataset: ImageNet. 
DL Model: VGG16.
XAI model: Deep Lift.
The primary objective has been achieved: having generated an adversarial image ($x_{adv}$), virtually identical to the original ($x$), the explanation map ($g$) of the adversarial image ($x_{adv}$) is now, incorrectly, that of the target image ($x_{target}$); essentially, the two rightmost columns are identical.}
\label{fig:imagenet-xai-vgg16}
\end{figure}

Figure~\ref{fig:imagenet-xai-vgg16} shows a sample result.

This work demonstrated how focused, undetectable modifications to the input data can result in arbitrary and significant adjustments to the explanation map. We showed that explanation maps of several known explanation algorithms may be modified at will. Importantly, this is feasible with a black-box approach, while maintaining the output of the model. We tested AttaXAI against the ImageNet and CIFAR100 datasets using 4 different network models.

\subsection{Patch of Invisibility: Naturalistic Black-Box Adversarial Attacks on Object Detectors \cite{Lapid2023}}
\label{sec:patch}
The implications of adversarial attacks can be far-reaching, as they can compromise the security and accuracy of systems that rely on DL. For instance, an adversarial attack on a vehicle-mounted, image-recognition system could cause it to misidentify a stop sign as a speed-limit sign \cite{eykholt2018robust}, potentially causing the vehicle to crash. As DL becomes increasingly ubiquitous, the need to mitigate adversarial attacks becomes more pressing. Therefore, research into adversarial attacks and defenses is a rapidly growing area, with researchers working on developing robust and secure  models that are less susceptible to such attacks.

\begin{figure}
\centering
\includegraphics[width=0.95\textwidth]{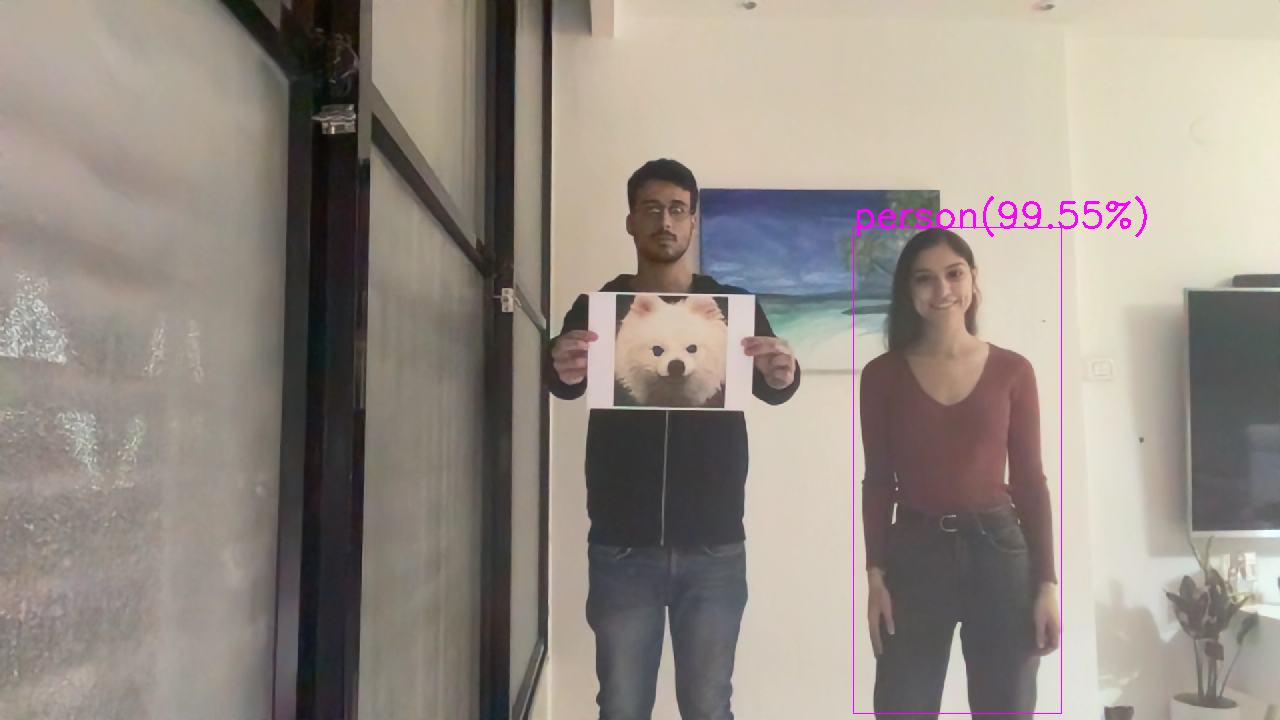}
\caption{An adversarial patch evolved by our gradient-free algorithm, which conceals people from an object detector.}
\label{fig:patch}
\end{figure}

In \cite{Lapid2023} we focused on fooling surveillance cameras (both indoor and outdoor), because of their ubiquity and susceptibility to attack, by creating adversarial patches (Figure~\ref{fig:patch}).

Our objective was to generate \textit{physically plausible} adversarial patches, which are performant and appear \textit{realistic}---\textit{without the use of gradients}.
An adversarial patch is a specific type of attack, where an image is modified by adding a small, local pattern that engenders missclassification. The goal of such an attack is to intentionally mislead a model into making an incorrect prediction or decision.

By ``physically plausible'' we mean patches that not only work digitally, but also in the physical world, e.g., when printed---and used. The space of possible adversarial patches is huge, and with the aim of reducing it to afford a successful search process, we chose to use pretrained generative adversarial network (GAN) generators.

Given a pretrained generator, we seek an input \textit{latent vector}, corresponding to a generated image that leads the object detector to err. We leverage the latent space's (relatively) small dimension, approximating the gradients using an Evolution Strategy algorithm \cite{wierstra2014natural}, repeatedly updating the input latent vector by querying the target object detector until an appropriate adversarial patch is discovered. 

\begin{figure}
\centering
\includegraphics[width=0.95\textwidth]{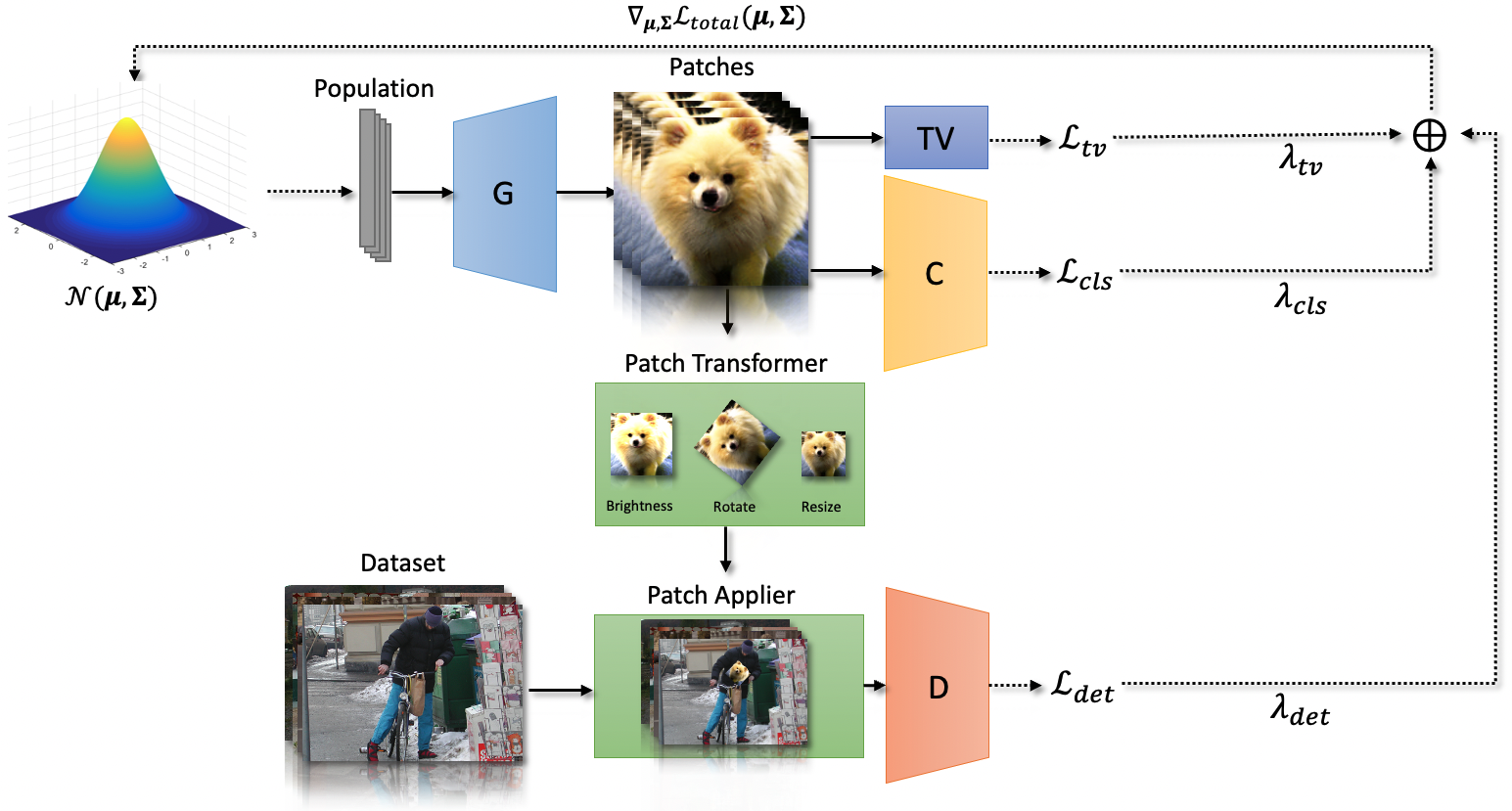}
\caption{Naturalistic Black-Box Adversarial Attack: Overview of framework. The system creates patches for object detectors by using the learned image manifold of a pretrained GAN ($G$) on real-world images (as is often the case, we use the GAN's generator, but do not need the discriminator). We use a pretrained classifier ($C$) to force the optimizer to find a patch that resembles a specific class, the $TV$ component in order to make the images as smooth as possible, and the detector ($D$) for the actual detection loss. Efficient sampling of the GAN images via an iterative evolution strategy ultimately generates the final patch.}
\label{fig:patch-alg}
\end{figure}

Figure~\ref{fig:patch-alg} depicts a general view of our approach. We search for an input latent vector that, given a pretrained generator, corresponds to a generated image that ``hides'' a person from the object detector.

The patches we generated can be printed and used in the real world. We compared different deep models and concluded that is is possible to generate patches that fool object detectors. The real-world tests of the printed patches demonstrated their efficacy in ``concealing'' persons, evidencing a basic threat to security systems. 

\clearpage
\section{Concluding Remark}
Our main conclusion from the works presented above is simple:
\begin{quote}
    When combined judiciously, EC and ML/DL reinforce each other to form a powerful alliance.
\end{quote}
\noindent
And we are fervently expanding this lineup of successful joint ventures...

\bibliography{bibliography}
\bibliographystyle{unsrtnat}

\end{document}